\documentclass[letterpaper, 10 pt, conference]{ieeeconf}  

\IEEEoverridecommandlockouts                              

\usepackage{graphicx, subcaption, booktabs, multirow, makecell,wrapfig}
\usepackage{dblfloatfix}
\usepackage{amsmath} 
\usepackage{amssymb}  
\usepackage{array}
\usepackage{tabularx}
\usepackage[table,xcdraw]{xcolor}
\usepackage{bm}
\usepackage{booktabs}

\newenvironment{flushitemize}{%
\begin{list}{$\bullet$}
   {\setlength{\leftmargin}{15pt}}%
    \setlength{\labelwidth}{20pt}
    \setlength{\itemindent}{0pt}
    \setlength{\labelsep}{0.5em}
 \setlength{\itemsep}{1pt}
 \setlength{\parskip}{0pt} 
 \setlength{\parsep}{0pt}}
 {\end{list}}

\title{\LARGE \bf
CAGE: Context-Aware Grasping Engine
}

\author{Weiyu Liu, Angel Daruna, Sonia Chernova
\thanks{Georgia Institute of Technology, Atlanta, Georgia, United States.
        {\tt\small wliu88,adaruna3,chernova@gatech.edu} }%
}

\begin{document}

\maketitle
\thispagestyle{empty}
\pagestyle{empty}

\begin{abstract}
Semantic grasping is the problem of selecting stable grasps that are functionally suitable for specific object manipulation tasks. In order for robots to effectively perform object manipulation, a broad sense of contexts, including object and task constraints, needs to be accounted for. We introduce the Context-Aware Grasping Engine, which combines a novel semantic representation of grasp contexts with a neural network structure based on the Wide \& Deep model, capable of capturing complex reasoning patterns.
We quantitatively validate our approach against three prior methods on a novel dataset consisting of 14,000 semantic grasps for 44 objects, 7 tasks, and 6 different object states. Our approach outperformed all baselines by statistically significant margins, producing new insights into the importance of balancing memorization and generalization of contexts for semantic grasping. We further demonstrate the effectiveness of our approach on robot experiments in which the presented model successfully achieved 31 of 32 suitable grasps. The code and data are available at: https://github.com/wliu88/rail\_semantic\_grasping
\end{abstract}

\section{Introduction}
Many methods have been developed to facilitate stable grasping of objects for robots \cite{lenz2015deep,mahler2017dex,levine2018learning,pinto2016supersizing}, some with optimality guarantees \cite{bicchi2000robotic}. However, stability is only the first step towards successful object manipulation. A grasp choice should also be based on a broader sense of context, such as object constraints (e.g., shape, material, and function) and task constraints (e.g., force and mobility) \cite{cini2019choice}. Considering all these factors allows humans to grasp objects intelligently, such as grasping the closed blades of scissors when handing them to another person. In robotics, \textit{semantic grasping} is the problem of selecting stable grasps that are functionally suitable for specific object manipulation tasks \cite{dang2012semantic}.

Most works on semantic grasping ground task-specific grasps to low-level features. Such features include object convexity \cite{song2010learning}, tactile data \cite{dang2012semantic}, and visual features \cite{hjelm2015learning}.
Leveraging similarities in low-level features allows grasps to be generalized between similar objects (e.g., from grasping a short round cup for pouring water to grasping a large tall cup for the same task \cite{dang2012semantic}). However, discovering structures in high-dimensional and irregular low-level features is hard and has prevented these methods from generalizing to a wider range of objects and tasks.

As a supplement to other types of information, such as geometrical or topological data, semantic data has been shown to improve robot reasoning for many applications, including task planning \cite{beetz2010towards}, plan repair \cite{boteanu2015towards}, and semantic localization \cite{pronobis2012large}. Semantic information has also been used for semantic grasping, in the form of hand written rules. For example, in \cite{antanas2018semantic}, the suitable grasp for handing over a cup is defined to be on the middle of the body of the cup. However, manually specifying every rule for different objects and tasks is not scalable, and limits adaptation to new situations.
%
\begin{figure}[t]
  \includegraphics[width=.8\linewidth]{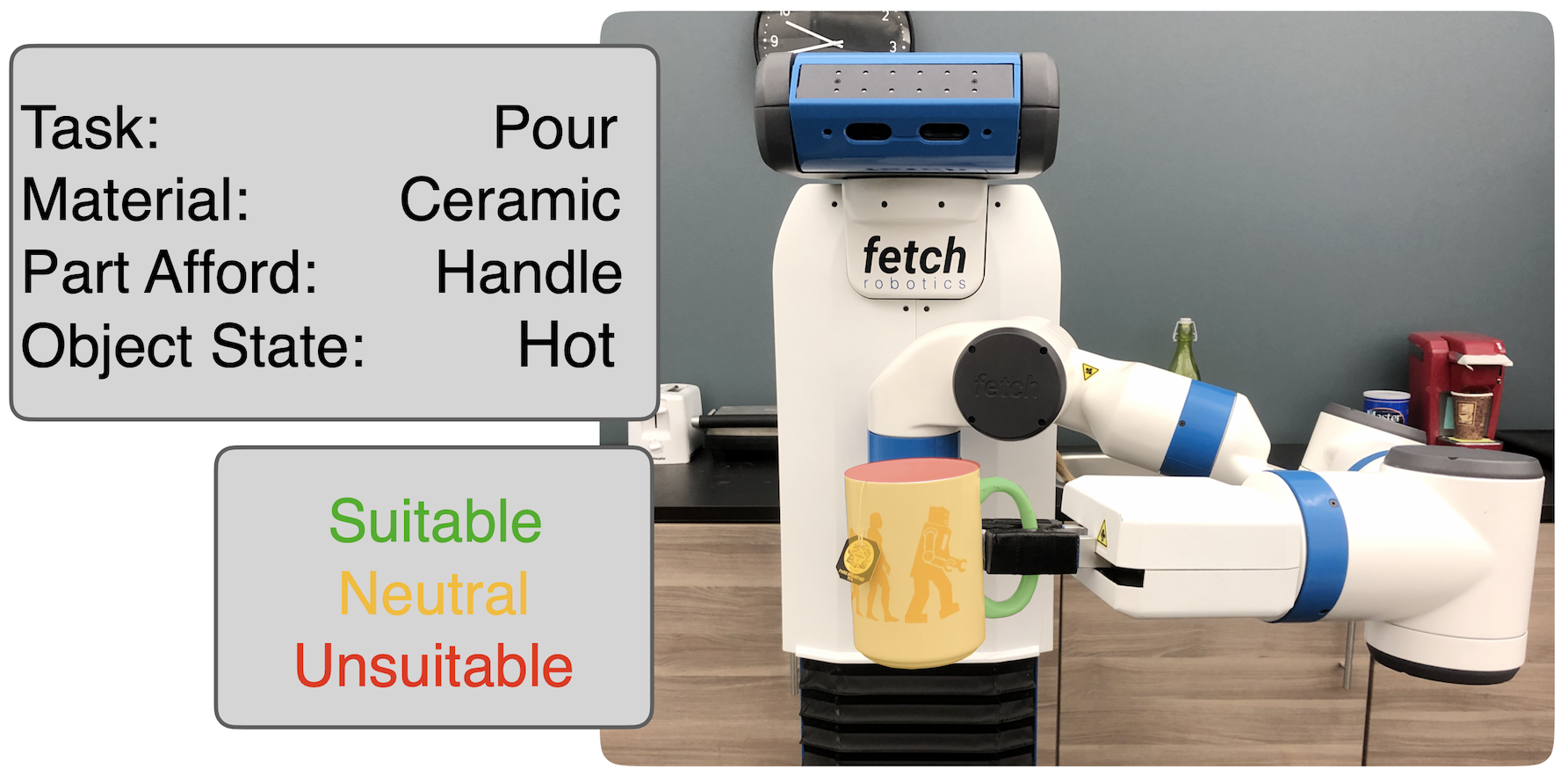}
  \centering
  \caption{The robot grasps the cup based on the context.}
  \label{fig:main_fig}
  \vspace{-0.5cm}
\end{figure}
In this work, we introduce \textbf{Context Aware Grasping Engine (CAGE)} to address semantic grasping by reasoning about grasp contexts, as shown in Figure 1. We first use perception modules, including pixel-wise affordance detection \cite{do2018affordancenet} and material classification from spectroscopic data \cite{erickson2019classification}, to acquire a semantic representation of the contexts. We then apply a reasoning module, based on the Wide \& Deep model \cite{cheng2016wide}, to infer the relations between extracted semantic features and semantic grasps. The perception and reasoning components together are capable of capturing the complex reasoning patterns for selecting suitable grasps and also generalizing to novel grasping situations.
We make the following contributions:  
\begin{flushitemize}
    \item We introduce a novel semantic representation that incorporates affordance, material, object state, and task to inform grasp selection and promote generalization.
    \item We apply a neural network structure, based on the Wide \& Deep model, to learn semantic grasps from data.
    \item We contribute a unique grasping dataset, SG14000, consisting of 14,000 semantic grasps for 44 objects, 7 tasks, and 6 different object states.
\end{flushitemize}
We quantitatively validate our approach against three prior methods on the above dataset. Our results show statistically significant improvements over existing approaches to semantic grasping. We further analyze how the semantic representation of objects facilitates generalization between object instances and object classes. Finally, we demonstrate how a mobile manipulator, the Fetch robot, can extract and reason about semantic information to execute semantically correct grasps on everyday objects.

\section{Related Work}
Below we discuss previous approaches to semantic grasping along with related works in affordance and material detection.
\begin{figure*}[t]
  \includegraphics[width=.9\linewidth]{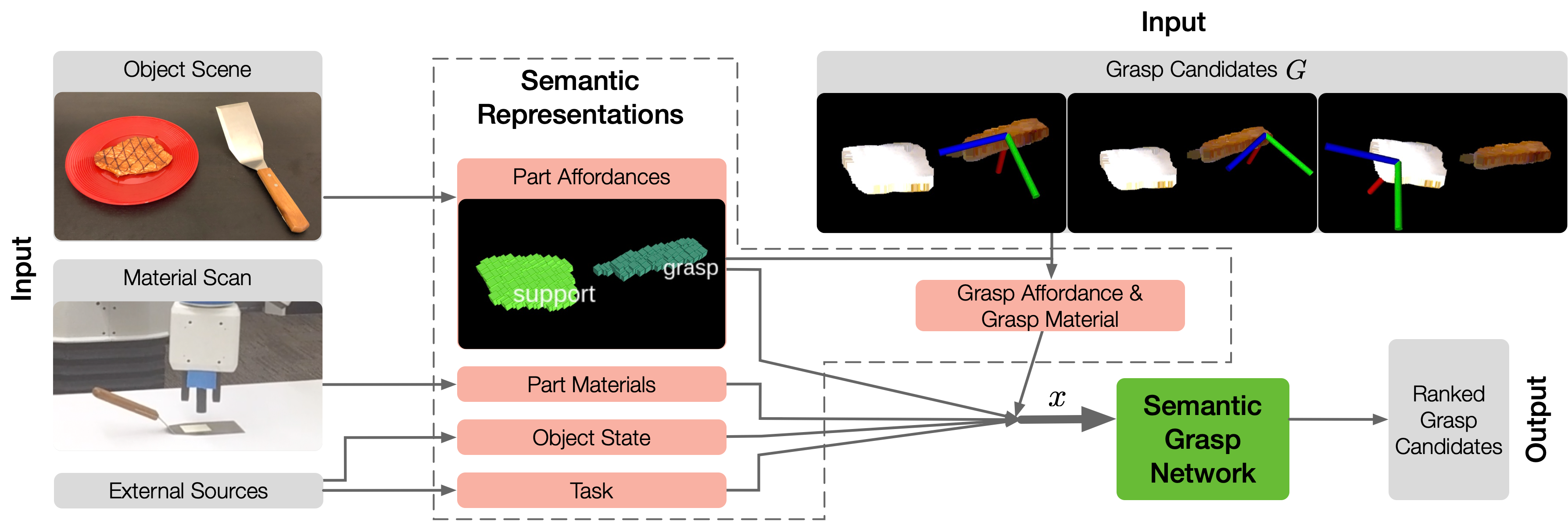}
  \centering
  \caption{The system overview of the Context-Aware Grasping Engine.}
  \label{fig:system}
  \vspace{-0.5cm}
\end{figure*}
\subsection{Semantic Grasping}
Multiple previous works have learned semantic grasps in a data driven manner, but they have relied on low-level or narrowly scoped features, making generalization challenging. Dang and Allen \cite{dang2012semantic} introduced an example-based approach which 
learns grasps appropriate for different tasks by storing visual and tactile data.
To improve generalization, Song et al. \cite{song2010learning} used Bayesian Networks to model the relations between objects, grasps and tasks; Hjelm et al. \cite{hjelm2015learning} learned a discriminative model based on visual features of objects. 
Instead of relying on low-level features, Aleotti et al. \cite{aleotti2011part} modeled semantic grasps with object parts obtained from topological shape decomposition and learned the parts that should be grasped at different stages of a task. 
Lakani et al. \cite{lakani2018exercising} learned the co-occurrence frequency of executive affordances (affordances associated with the tasks) and manipulative affordances (affordances of the parts which need to be grasped). In contrast, we are reasoning over a wider variety of contextual information. 

Other works have used more abstract semantic representations, but leveraged rule-based inference mechanisms that must be encoded manually. Antanas et al. \cite{antanas2018semantic} encoded suitable grasp regions based on probabilistic logic descriptions of tasks and segmented object parts. 
Works on object affordance detection \cite{lakani2019towards,chu2019learning,do2018affordancenet,nguyen2016detecting,nguyen2017object}
have leveraged the affordances of object parts to define the correspondences between affordances and grasp types (e.g., rim grasp for parts with \textit{contain} or \textit{scoop} affordance.) Detry et al. \cite{detry2017task} built on these works by training a separate detection model using synthetic data generated by predefined heuristics to detect suitable grasp regions for each task. Kokic et al. \cite{kokic2017affordance} assigned preferences to affordances in different tasks, which were then used to determine grasp regions to use or avoid. Instead of manually specifying semantic grasps based on semantic representations, we learn the relations between these semantic features and suitable grasps from data.

\subsection{Affordance Detection}
Affordance detection has been defined as the problem of pixel-wise labeling of object parts by functionality. Myers et al. \cite{myers2014affordance} first proposed to use hand-crafted features to detect affordances from 2.5D data. Nguyen et al. \cite{nguyen2016detecting} later learned useful features from data with an encoder-decoder structure. Building on these methods, Nguyen et al. \cite{nguyen2017object} and Do et al. \cite{do2018affordancenet} developed models to simultaneously perform object detection and affordance segmentation. Joint training on these two objectives helps their models to achieve state-of-art performance. Sawatzky et al. \cite{sawatzky2017weakly} tried to reduce the cost of labeling with weakly supervised learning. Chu et al. \cite{chu2019learning} explored transferring from systhetic data to real environment with unsupervised domain adaptation. We are the first to leverage part affordances to learn semantic grasping from data.

\subsection{Material Detection}
Material detection is the problem of the inferring an object's constituent materials from observations of the object. While most approaches seek to identify material classes from images \cite{schwartz2019recognizing, hu2011toward, bell2015material}, in more recent work, Erickson et al. \cite{erickson2019classification} used spectral data obtained from a hand-held spectrometer for material classification, achieving a validation accuracy of 94.6\%. 
Previous works in robotics \cite{lakshmiautonomous} have successfully used this approach for tool construction. We also leverage this approach to infer materials of objects.
%

\section{Problem Definition}
Given a set of grasps $G$ from any stable grasp sampling methods, our objective is to model the probability distribution $P(G|C)$, such that the probability $p(g|c)$ is maximized for a grasp $g \in G$ that is suitable for the context $c \in C$. In this work, we model contexts $C = \{O, T\}$ as consisting of information related to both objects, $O$, and the tasks, $T$. To avoid the computational complexity of directly modeling $P(G|C)$, we simplify the problem by learning a discriminative model for $P(Y|G,C)$, where $Y$ is a class label indicating whether grasps in $G$ are suitable for contexts in $C$. With this discriminative model, we can rank grasp candidates $g \in G$ for various contexts $C$, ensuring that more suitable grasp candidates are ranked higher than less suitable ones given a context, i.e., $p(y|g_i, c) > p(y|g_j, c),\: \forall g_i, g_j \in G$ if $g_i$ is more suitable than $g_j$ for the context $c$.

\section{Context-Aware Grasping Engine}
\label{sec:feature_ext}
To address the above problem, we present the Context-Aware Grasping Engine (CAGE) (Figure \ref{fig:system}), which takes as input the context $c$ and a set of grasp candidates $G$, and outputs a ranking of grasps ordered by their suitability to the context $c$. In this work, we model context $c$ as the label $t$ of the task being performed and the following information about object $o$: a point cloud of the object, a spectral reading of the object from the SCiO sensor for material recognition, and a label specifying the object state\footnote{In the current implementation, the task and object state are hand labeled.  In general, task information can be obtained from the robot's task plan.  Object state can be obtained through techniques such as \cite{thomason2016learning} or \cite{amiri2018multi}.}.
For each grasp $g \in G$, CAGE extracts semantic features from both the context inputs and the grasp, forming the vector $\bm{x}$.  In this formulation, $\bm{x}$ serves as a unified abstraction of $g$ and $c$, allowing us to model $P(Y|G,C)$ as $P(Y|X)$. Finally, we use $\bm{x}$ as input into the Semantic Grasp Network, which ranks grasps based on $p(y|\bm{x})$. 
In Section \ref{sec:representations}, we describe the process for extracting the semantic features $\bm{x}$. Then in Section \ref{sec:engine}, we present details of the semantic grasp network. 
%

\section{Semantic Representations} \label{sec:representations}
In this section, we introduce semantic representations of objects and grasps extracted from context inputs.

\subsection{Affordances of Object Parts} \label{sec:aff}
We use affordance of each object part as one of the semantic features to capture constraints on suitable grasp regions. Part affordances present an abstract representation to help generalize among object instances and classes (e.g. \textit{contain} affordance can be observed in most mugs, and also other objects such as bowls and pans). Compared to other topological decompositions of objects (e.g., curvature-based segmentation \cite{tenorth2013decomposing}), the focus on functionalities also makes part affordances more suitable for determining task-dependent grasps. 

We use the state of the art affordance detecton model, AffordanceNet \cite{do2018affordancenet}, to extract part affordances. AffordanceNet takes as input an RGB image of the scene and simultaneously outputs class labels, bounding boxes, and part affordances of the detected object. The affordance prediction is produced as a 2D segmentation mask; however, grasp candidates often correspond to 3D points. Therefore, we superpose the affordance segmentation mask with the object point cloud segmented from a plane model segmentation to map affordance labels from 2D pixels to 3D points. 
%

\subsection{Materials of Object Parts} \label{sec:mat}
In addition to affordances of object parts, we extract materials of object parts as another semantic feature to help rank grasp candidates. AffordanceNet and other affordance detection methods do not explicitly take material into account. Therefore, adding materials helps refine semantics of objects and facilitate more complex reasoning (e.g., grasping the part of knife with the \textit{cut} affordance is safe if the part is made of plastic).

In order to infer the materials of object parts, we follow the approach of \cite{erickson2019classification}. A handheld spectrometer, the SCiO sensor, is used by the robot to get spectral readings of objects at various poses. The neural network architecture in \cite{erickson2019classification} then classifies the material from the 331 dimensional vector of real values in each spectral scan. 

\vspace{-0.10cm}
\subsection{Grasps} \label{sec:grasp}
Given a set of grasp candidates, the semantic meaning of each grasp is determined based on the grasp's relation to the object. Specifically, for each grasp, we assign the affordance and material of the object part closest to the grasp as the semantic representation of the grasp, which we call grasp affordance and grasp material, respectively. Modeling these two semantic features has shown to be extremely effective in our experiments as many simple semantic grasping heuristics can be captured solely based on these two features. For example, grasps on a cup with the grasp affordance \textit{contain} are usually not suitable for the pouring task as these grasps obstruct the opening. 

In order to extract the grasp affordance and grasp material, we use a kd-Tree to efficiently search for the closest point on the object point cloud to the center of the grasp. Then the affordance and material corresponding to the part of object the point lies on are assigned to the grasp. 

\smallskip 
We combine the extracted semantic features above to create the unified semantic representation $\bm{x}$. Including task, object state, grasp affordance, grasp material, part affordances and part materials, $\bm{x}$ has a dimension of $4 + 2N $, where N is the number of parts. 

\vspace{-0.1cm}
\section{Semantic Grasp Network} \label{sec:engine}
To effectively reason about the semantics of contexts, a model needs both memorization and generalization capabilities. Memorization facilitates forming reasoning patterns based on combinations of semantic features (e.g., combining the \textit{contain} part affordance, \textit{cup} object, and \textit{pour} task to avoid grasping the opening part of a cup for pouring). Generalization helps unifying previously seen and new contexts (e.g., generalize avoiding grasps on the opening part of bottle for the \textit{pour} task to \textit{scoop} and \textit{handover}). Additionally, combining these two capabilities allows the model to refine generalized rules with exceptions (e.g., grasping the opening part of the cup for handover is fine if the cup is \textit{emtpy}).

In this work, we introduce a neural network based on the Wide \& Deep model \cite{cheng2016wide} to predict $p(y|\bm{x})$ given the extracted semantic features $\bm{x}$ of the context $c$ and the grasp $g$. The wide component of the model promotes memorization of feature combinations with a linear model, and the deep component aids generalization by learning dense embeddings of features. Below, we present details of the model.

\subsection{The Wide Component}

\begin{figure}[t]
  \includegraphics[width=\linewidth]{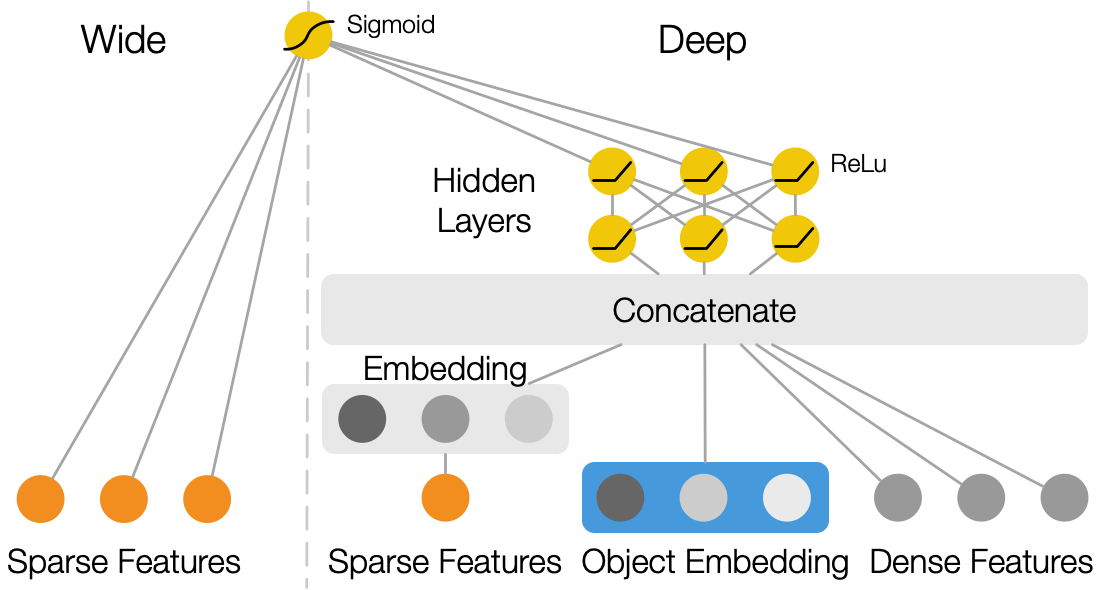}
  \centering
  \caption{Model structure of the semantic grasp network.}
  \label{fig:architecture}
  \vspace{-0.5cm}
\end{figure}

The wide component captures the frequent co-occurrence of semantic features and target labels. This part of the model takes in one-hot encodings of sparse features. Here, each sparse feature corresponds to a dimension in $\bm{x}$, i.e., a semantic feature.
As shown in Figure \ref{fig:architecture}, the wide component is a generalized linear model of the following form:
\begin{align}
    y = \bm{w}^T \bm{x}^{wide} + b
\end{align}
where y is the prediction, $\bm{x}^{wide}$ is the concatenation of one-hot encodings of all sparse features.

\subsection{The Deep Component}

The deep component achieves generalizations of feature combinations by learning a low-dimensional dense embedding vector for each feature. For example, despite never encountering the pair of the \textit{pound} task and the \textit{support} affordance before, the network can still generalize to this new combination because the embedding vectors of both elements have been paired and trained individually with other affordances and tasks. 


The deep component is a feed-forward neural network, as shown in Figure \ref{fig:architecture}. Sparse features are first converted into dense embedding vectors. The embedding vectors are intialized randomly and the values are learned through training. The embedding vectors, along with object embedding (see Sec.~\ref{sec:obj_embedding}) and other dense features (e.g., visual features) are combined to create the input $\bm{x}^{deep}$ into the hidden layers of the neural network. Each hidden layer has the following form:
\begin{align}
    \bm{a}^{(l+1)}=f({\bm{w}^{(l)}}^T\bm{a}^{(l)} + \bm{b}^{(l)})
\end{align}
where $l$ is the layer number, $f$ is the activation function, $\bm{w}^{(l)}$ and $\bm{b}^{(l)}$ are the neural network parameters, and $\bm{a}^{(l)}$ is the activation. The first activation is the input, i.e., $\bm{a}^{(0)} = \bm{x}^{deep}$.

\subsection{Object Embedding}
\label{sec:obj_embedding}
In this section, we discuss in detail how to create the object embedding that encodes semantic information of object parts. A naive approach is to simply concatenate embeddings of all parts.  However, this approach can lead to features that are difficult to generalize because objects consist of different numbers of parts and a canonical order of parts for concatenation is undetermined. 

Inspired by the \textit{propagation} and \textit{pooling} functions in Graph Neural Networks \cite{zhou2018graph}, we create a vector representation of objects that handles arbitrary numbers of parts and is invariant to their orderings.
Specifically, the semantic features of each part, including part affordance and part material, are first mapped to dense embedding vectors and then together passed through a \textit{propagation} function, creating an embedding for each part $\bm{v}_{p_i}$. All part embeddings are then combined through a \textit{pooling} function (we use average pooling) to create the object embedding $\bm{v}_{o}$. Formally,

\noindent\begin{minipage}{.5\linewidth}
    \vspace{-0.2cm}
    \begin{align}
      \bm{v}_{p_i} &= f(\bm{w}^T \bm{x}_{p_i} + \bm{b})
    \end{align}
    \vspace{0.01cm}
\end{minipage}%
\begin{minipage}{.5\linewidth}
    \vspace{-0.2cm}
    \begin{align}
      \bm{v}_{o} &= \dfrac{1}{N} \sum_{i=1}^{N} \bm{v}_{p_i}
    \end{align}
    \vspace{0.1cm}
\end{minipage}

\noindent where $\bm{x}_{p_i}$ represents the concatenated embeddings of a part's affordance and material, $N$ is the number of parts for the object, $f$ is the activation function, and $\bm{w}$ and $\bm{b}$ are the parameters of the propagation function.





\subsection{Joint Training of Wide and Deep Components}

To combine the wide and deep components, we compute a weighted sum of the outputs of both components. The sum is then fed into a softmax function to predict the probability that a given grasp has label $y$. For a multi-class prediction problem, the model's prediction is:
\begin{align}
    P(y|\bm{x}) = \text{softmax}(\bm{w}_{wide}^T \bm{x}^{wide} + {\bm{w}_{deep}}^T \bm{a}^{(l_f)} + b)
\end{align}
where $\bm{w}_{wide}$ and $\bm{w}_{deep}$ are the model parameters of the wide and the deep components, $b$ is the bias term, and $\bm{a}^{(l_f)}$ is the final activations of the deep component.

The combined model is trained end to end with the training set of contexts $C_{train}$ and grasps $G_{train}$, from which semantic features $X_{train}$ are extracted, as well as corresponding ground-truth labels $Y_{train}$. Our training objective is to minimize the negative log-likelihood of the training data. We used Adam \cite{kingma2014adam} for optimization with default parameters (learning rate=$1e^{-3}$, $\beta_1$=$0.9$, $\beta_2$=$0.999$, $\epsilon$=$1e^{-8}$). We trained the models fully to 150 epochs.


\begin{table}[b]
\centering
\begin{tabular}{ll}  
\toprule
Object classes & cup, spatula, bowl, pan, bottle \\
Materials & plastic, metal, ceramic, glass, stone, paper, wood \\
Tasks & pour, scoop, poke, cut, lift, hammer, handover \\
States & hot, cold, empty, filled, lid on, lid off \\
\bottomrule
\end{tabular}
\caption{SG14000 dataset used to validate CAGE}
\label{tas}
\vspace{-0.3cm}
\end{table}

\begin{figure}[b]
  \includegraphics[width=0.8\linewidth]{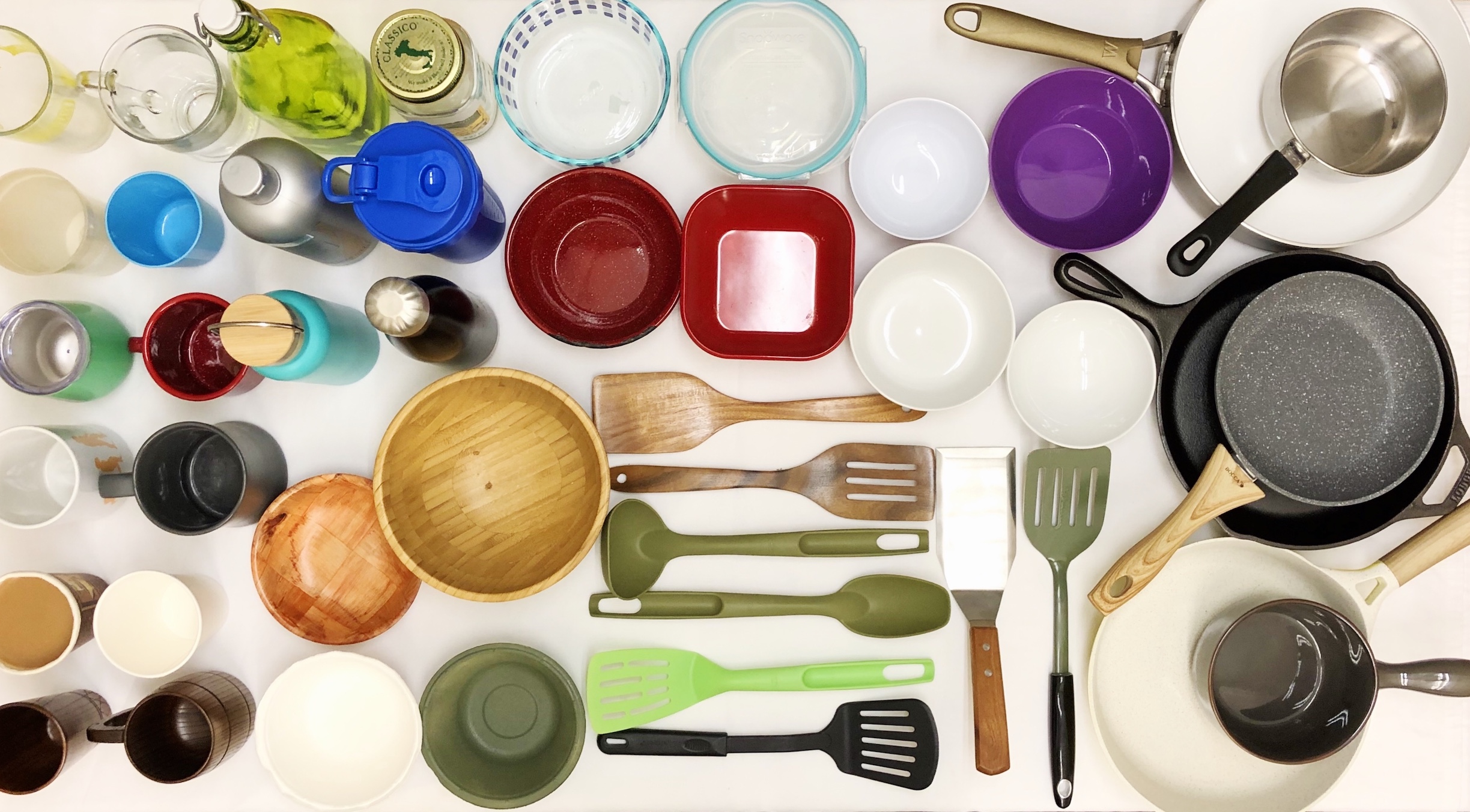}
  \centering
  \caption{44 objects in the SG14000 dataset.}
  \label{fig:objects}
  \vspace{-0.5cm}
\end{figure}

\section{Experimental Setup}
In this section we discuss the details of our novel semantic grasping dataset that is used to validate CAGE, the baseline approaches, and the evaluation metric.

\begin{figure*}[h]
    \centering
    \begin{subfigure}[t]{0.32\textwidth}
        \includegraphics[width=\textwidth]{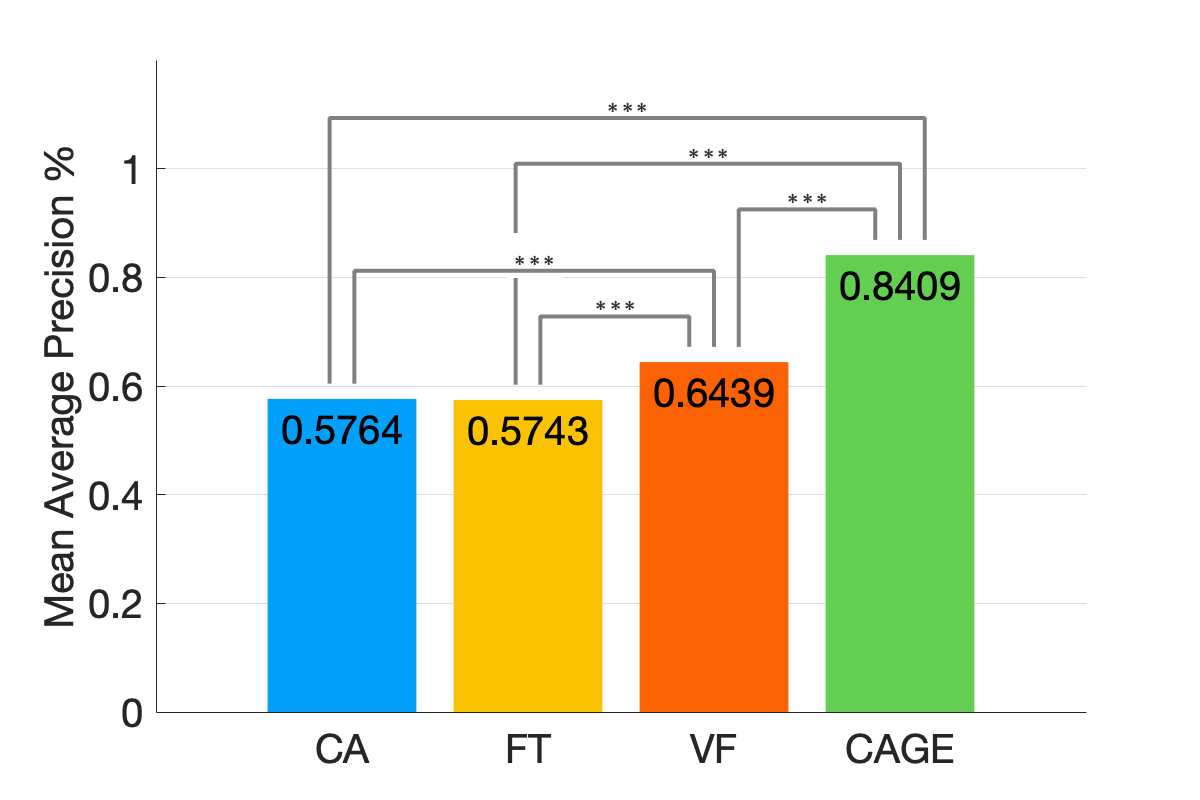}
        \centering
        \caption{Context-aware grasping}
        \label{fig:exp4}
    \end{subfigure}%
    ~ 
    \begin{subfigure}[t]{0.32\textwidth}
        \includegraphics[width=\textwidth]{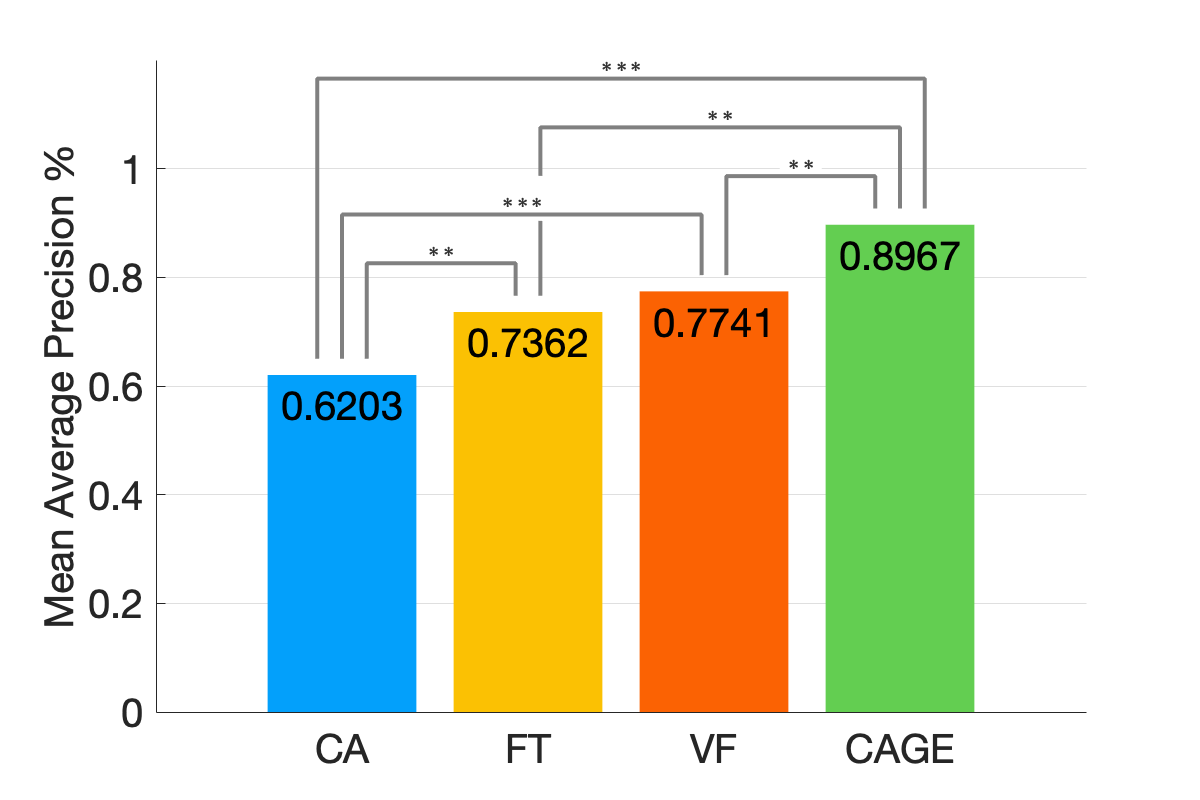}
        \centering
        \caption{Object instance generalization}
        \label{fig:exp1}
    \end{subfigure}
    ~
    \begin{subfigure}[t]{0.32\textwidth}
        \includegraphics[width=\textwidth]{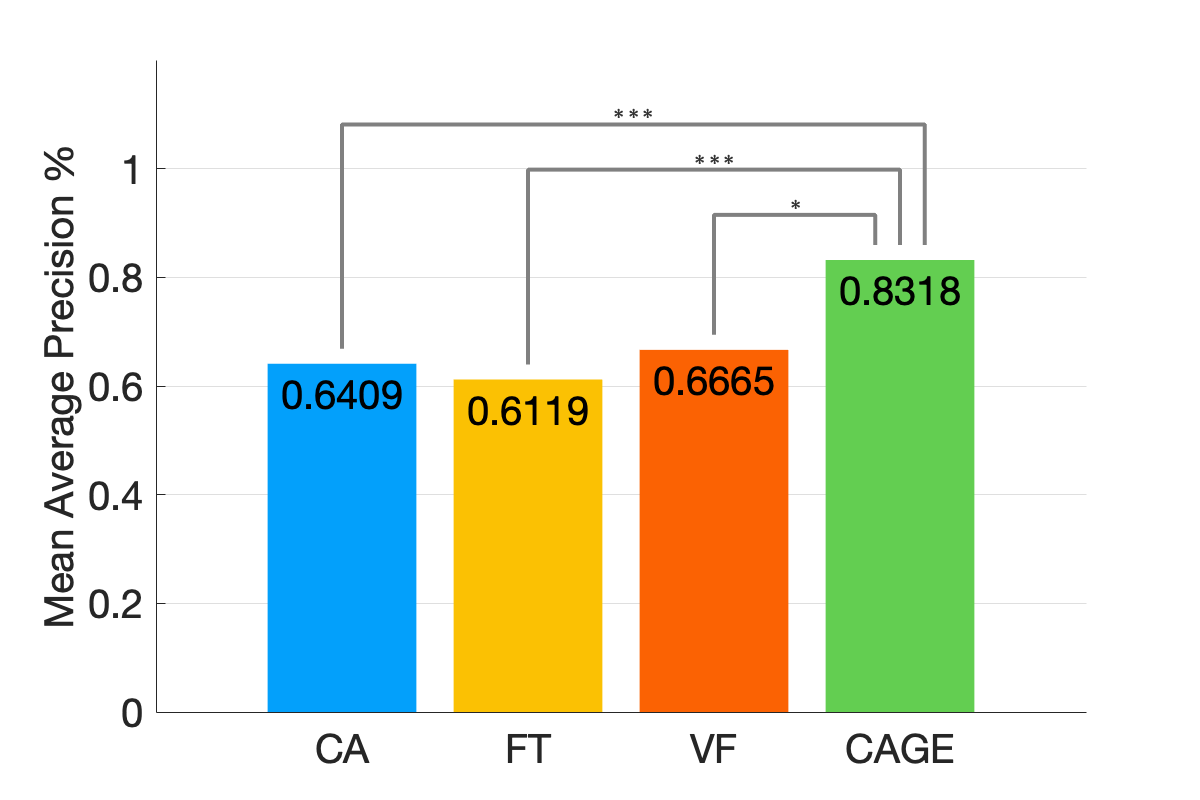}
        \centering
        \caption{Object class generalization}
        \label{fig:exp2}
    \end{subfigure}
    \caption{MAP\% of CAGE and baseline methods validated on the SG14000 dataset.}
    \vspace{-0.5cm}
\end{figure*}

\subsection{SG14000 Dataset}

In order to establish a benchmark dataset for semantic grasping, we collected a dataset of 14,000 semantic grasps for 700 diverse contexts. A summary of the dataset is shown in Table \ref{tas}. As shown in Figure \ref{fig:objects}, we chose objects with a variety of materials, shapes, and appearances to reflect the diversity of objects robots might encounter in real environments. For each object, we first used the Antipodal grasp sampler \cite{agile} to sample 20 stable grasp candidates. Then we labeled all grasps as \textit{suitable}, \textit{neutral}, or \textit{not suitable} 
for a given context. Using three class labels \cite{detry2017task} allows methods to directly model positive and negative preferences (e.g., for handing over a cup filled with hot tea, grasping the body is preferred and grasping the opening should be avoided). 


\subsection{Baselines}
We compare \textit{\textbf{CAGE}} to the following three baselines that are representative of existing works:

\begin{flushitemize}
\item \textit{\textbf{Context Agnostic (CA)}} represents grasping strategies \cite{agile,mahler2017dex,levine2018learning} that focus on grasp stability and ignore semantic context. Given a set of stable grasp candidates, this method ranks them randomly.   

\item \textit{\textbf{Classification of Visual Features (VF)}} \cite{hjelm2015learning} performs semantic grasping based on low-level visual features. The features include object descriptors that represent object shape, elongatedness, and volume, visual descriptors such as image intensity and image gradients, grasp features that encode grasping position and orientation. This method ranks grasp candidates based on prediction scores. 

\item \textit{\textbf{Frequency Table of Affordances (FT)}} performs semantic grasping based on detected part affordances. This benchmark is adapted from \cite{lakani2018exercising}, with the difference that our representation replaces executive affordances with contexts. This method learns the co-occurrence frequency of grasp affordances and contexts. Grasp candidates with the grasp affordances that are more frequently used in the given context are ranked higher.
\end{flushitemize}

We do not benchmark against methods that manually define semantic grasps \cite{antanas2018semantic,kokic2017affordance} because our dataset contains 700 unique contexts.

\subsection{Evaluation Metric}
We use Average Precision (AP) for evaluating the performance of all methods. As a single number that summarises the Precision-Recall curve, AP is widely used for assessing ranking systems. An AP score of $1.0$ indicates that all \textit{suitable} grasps are ranked higher than all neutral and unsuitable grasps. If no \textit{suitable} grasps exists for a context, we evaluate whether \textit{neutral} grasps are ranked higher than \textit{not suitable} grasps.
Mean Average Precision (MAP) is reported.  

\section{Experiments}

In this section, we compare our method to the baselines on the SG14000 dataset. We formulate three evaluation tasks for assessing each method, evaluating the ability to select semantic grasps in different contexts, generalize between object instances, and generalize between object classes\footnote{Where ever data is split into training and test sets in an evaluation task, we report results averaged over 10 random splits. Statistical significance was determined using a paired t-test, with each split treated as paired data *, **, and *** indicate $p<0.05, 0.01$ and $0.001$, respectively.}. 
We demonstrate the effectiveness of our method on a real robot. 



\subsection{Context-Aware Grasping}
In this evaluation task, we examine each method's ability to successfully rank suitable grasps for different contexts. All contexts are split into 70\% training and 30\% testing. 

As shown in Figure \ref{fig:exp4}, \textbf{\textit{CAGE}} outperformed all baselines by statistically significant margins. This result highlights our model's ability to collectively reason about the contextual information of grasps. \textbf{\textit{VF}} also performed better than the other two baselines, and demonstrated that visual features also help generalize the similarities observed in semantic grasps. However, the significantly better performance of \textbf{\textit{CAGE}} shows that our representation of context can promote generalization not only at the visual level, but also at the semantic level. One of possible reasons that \textbf{\textit{FT}} performed worse than \textbf{\textit{CA}} is that it failed to distinguish different contexts and overfitted to the bias in training data. This result again confirms that part affordances need to be treated as part of a broader set of semantic features that together determine the suitability of grasps.

\begin{wraptable}{l}{0.22\textwidth}
    \vspace{-0.4cm}
    \begin{tabular}{ll}  
        \toprule
        Model & MAP\% \\
        \midrule
        \textbf{Wide and Deep} & 0.8409 \\
        Without Deep & 0.7657 \\
        Without Wide & 0.8275 \\
        Without States & 0.8006 \\
        Without Tasks & 0.7295 \\
        \bottomrule
    \end{tabular}
    \caption{Ablation}
    \label{tab:ablation}
    \vspace{-0.3cm}
\end{wraptable}

To gain deeper insights into our proposed model and the effect of different contextual information, we performed an ablation study on our model. As shown in Table \ref{tab:ablation}, removing either the \textit{\textbf{Deep}} or the \textit{\textbf{Wide}} component of the model hinders the model's ability to generalize different contexts. The ablation also shows that removing task information (e.g., \textit{pour}) has a negative impact on the performance because different tasks introduce drastically different constraints on grasp suitability. Additionally, the ablation shows that reasoning about object states (e.g. \textit{hot}, \textit{full}), which is lacking in prior work, is also important for accurately modeling semantic grasps.


\subsection{Object Instance Generalization}
In this evaluation task, we investigate whether each method can generalize semantic grasps between object instances from the same class. For each task and each object class, we train on 70\% of object instances and test on the remaining 30\%.
Figure \ref{fig:exp1} presents the results. \textit{\textbf{CAGE}} again outperformed all other baselines. Note that since the training and test sets share the same object class and task, less contextual information is required by the models.  
However, the improvement of \textit{\textbf{CAGE}} over \textit{\textbf{VF}} in this experiment reaffirms the importance of affordances of object parts as a representation for modeling objects and semantic grasps.

\subsection{Object Class Generalization}
In this evaluation task, we test the generalization of semantic grasps between object classes. We split the dataset by task; for each task, we set aside data from one object class for testing and use remaining object classes for training.  

As shown in Figure \ref{fig:exp2}, only \textit{\textbf{CAGE}} has statistically significant differences in performance from the other methods. The similar performance of \textit{\textbf{VF}} and \textit{\textbf{CA}} might be explained by the diverse set of objects in our data. Since objects in different categories have drastically different appearances, generalization of semantic grasps based on visual similarities is hard to achieve. \textit{\textbf{FT}} also performed poorly on this experiment because it cannot distinguish different object classes, while different object classes can have distinct grasps suitable for the same task.

\subsection{Robot Experiment}
\begin{figure}[t]
  \includegraphics[width=\linewidth]{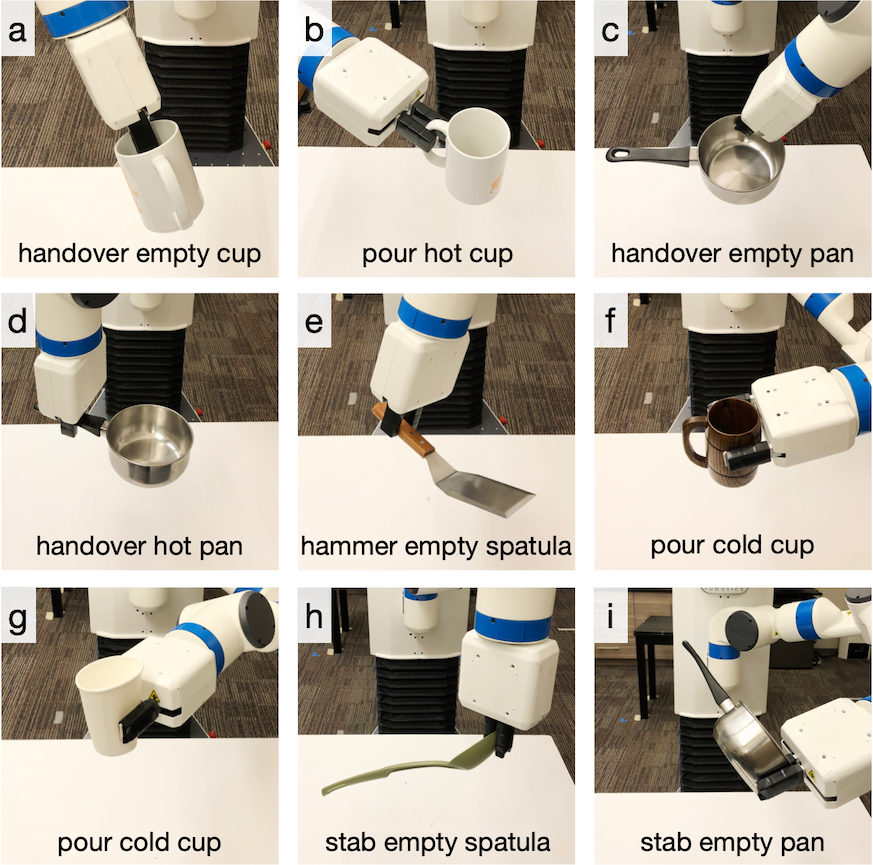}
  \centering
  \caption{Examples of semantic grasps executed by the robot.}
  \label{fig:qualitative_results}
  \vspace{-0.5cm}
\end{figure}

In our final experiment, we evaluated the effectiveness of our method on a Fetch robot \cite{wise2016fetch} equipped with a RGBD camera, a 7-DOF arm, and a parallel-jaw gripper. For testing, we randomly withheld 32 contexts from our dataset, of which 16 were semantically feasible and 16 were not, covering 14 different objects and all 7 tasks.  The objective of this experiment was to validate \textit{\textbf{CAGE}}'s success rate in selecting semantically meaningful grasps for feasible cases, and its ability to reject grasps in semantically infeasible cases (e.g., \textit{hammer} with a \textit{ceramic bowl}, and \textit{scoop} with a \textit{flat spatula}).

We trained a \textit{\textbf{CAGE}} model with the remaining data and tested grasping with each of the withheld contexts on the robot. The test for each context proceeded as follows: We placed the object defined in the context on the table and 50 grasp candidates were automatically sampled from the RGBD data with the Antipodal grasp sampler \cite{agile}. \textit{\textbf{CAGE}} extracted semantic information from the object and each grasp candidate, and then used the trained network to predict a ranking of all grasp candidates. If the trained network predicted that all grasps have suitability probabilities below a threshold (0.01 was used in the experiment), the robot rejected all grasp candidates. Otherwise, the robot executed the first grasp candidate from the ranked list. 

In the experiment, 15/16 (94\%) semantically feasible contexts had successful grasps (stable grasp lifting object off the table). While 1 grasp failed due to a bad grasp sample (i.e. wrap grasp on a pan), 16/16 (100\%) grasps were correct for the context. The remaining 16/16 (100\%) semantically infeasible contexts were correctly predicted by the model to have zero suitable grasps. The material classification incorrectly predicted \textit{metal} as \textit{glass} once; however, the misclassification did not cause an error in the ranking of semantic grasps.

Shown in Figure \ref{fig:qualitative_results} are examples of semantic grasps executed by the robot. With \textit{\textbf{CAGE}}, the robot was able to select suitable grasps considering task (a and b), state (c and d), and material (e). Generalization of semantic grasps between object instances (f and g) and classes (h and i) was also achieved, due to the abstract semantic representations and the semantic grasp network.

\section{CONCLUSIONS}

This work addresses the problem of semantic grasping. We introduced a novel semantic representation that incorporates affordance, material, object state, and task to inform grasp selection and promote generalization. We also applied the Wide \& Deep model to learn suitable grasps from data. Our approach results in statistically significant improvements over existing methods evaluated on the dataset of 14,000 semantic grasps for 44 objects, 7 tasks, and 6 different object states. An experiment on the Fetch robot demonstrated the effectiveness of our approach for semantic grasping on everyday objects.




%
%
%


\section*{ Acknowledgments}
This work is supported in part by NSF IIS 1564080, NSF GRFP DGE-1650044, and ONR N000141612835.

\bibliographystyle{IEEEtran}
\bibliography{mybibfile}

\end{document}